# Semi-Supervised Prediction of Gene Regulatory Networks Using Machine Learning Algorithms

Nihir Patel[1] and Jason T. L. Wang[1,2,*]

[1]Bioinformatics Program, New Jersey Institute of Technology, Newark, NJ 07102, USA

[2]Computer Science Department, New Jersey Institute of Technology, Newark, NJ 07102, USA

*Corresponding author: wangj@njit.edu

## Abstract

Use of computational methods to predict gene regulatory networks (GRNs) from gene expression data is a challenging task. Many studies have been conducted using unsupervised methods to fulfill the task; however, such methods usually yield low prediction accuracies due to the lack of training data. In this article, we propose semi-supervised methods for GRN prediction by utilizing two machine learning algorithms, namely support vector machines (SVM) and random forests (RF). The semi-supervised methods make use of unlabeled data for training. We investigate inductive and transductive learning approaches, both of which adopt an iterative procedure to obtain reliable negative training data from the unlabeled data. We then apply our semi-supervised methods to gene expression data of *Escherichia coli* and *Saccharomyces cerevisiae*, and evaluate the performance of our methods using the expression data. Our analysis indicated that the transductive learning approach outperformed the inductive learning approach for both organisms. However, there was no conclusive difference identified in the performance of SVM and RF. Experimental results also showed that the proposed semi-supervised methods performed better than existing supervised methods for both organisms.

# 1. Introduction

## 1.1. Background

Using gene expression data to infer gene regulatory networks (GRNs) is a key approach to understand the relationships between transcription factors (TFs) and target genes that may aid to uncover underneath biochemical pathways governed by the TFs. Analyzing individual TF and gene associations to the level that induces biological significance through wet-lab experiments is a practically challenging, costly and time-consuming task (Pe'er & Hacohen, 2011). It is therefore useful to adopt computational methods to obtain similar information, because results obtained from such methods can easily and quickly be reproduced through inexpensive procedures while allowing multiple methods to explore data to validate outcomes (Cerulo et al., 2010; Gillani et al., 2014; Lingeman & Shasha, 2012; Maetschke et al., 2014).

Various computational methods for performing unsupervised, supervised and semi-supervised prediction of GRNs have been proposed. These methods employ a variety of techniques ranging from Boolean networks (Lähdesmäki et al., 2003) and Bayesian networks (Acerbi et al., 2014; Vignes et al., 2011) to compressive sensing (Chang et al., 2014). Integrated toolkits combining different network inference methods are also available (Hurley et al., 2015). Many of the methods are unsupervised. In previous studies, several authors have shown that supervised and semi-supervised methods outperformed unsupervised methods (Cerulo et al., 2010; Maetschke et al., 2014). However, supervised methods require training data to contain both positive and negative examples, which are difficult to obtain. In contrast, semi-supervised methods can work with a large number of unlabeled examples, which are much easier to obtain (Elkan & Noto, 2008). In this paper, we propose new semi-supervised methods capable of predicting TF-gene interactions in the presence of unlabeled training examples.

In order to predict GRNs it is essential to understand possible TF-gene interaction types. Figure 1 illustrates a gene regulatory network between the *Escherichia coli* transcription factor FNR and several target genes. The network is created with Cytoscape (Shannon et al., 2003) using true TF-gene interactions obtained from RegulonDB (Salgado et al., 2013). In general, if there exists an experimentally verified interaction between a TF and a target gene, then such an interaction is considered to be known. The known interactions are generated through wet-lab or sometimes dry-lab experiments that are indirectly associated with wet-lab experiments, and curated based on experimental outputs. On the contrary, TF-gene interactios that are not yet experimentally verified are considered to be unknown. In Figure 1, solid edges represent known interactions and dotted edges represent unknown interactions. There are two types of known interactions: up-regulation and down-regulation. If a transcription factor activates a gene, then the gene is up-regulated. If a transcription factor inhibits (or represses) a gene, then the gene is down-regulated.

## 1.2. Network prediction

There are three types of computational methods for predicting gene regulatory networks (GRNs); they are supervised, semi-supervised, and unsupervised methods (Maetschke et al., 2014). The first two types of methods differ primarily based on whether training examples are labeled. In supervised methods, each training example must have a (positive or negative) label. In semi-supervised methods, some training examples are labeled and some are unlabeled. The training examples for the supervised and semi-supervised methods are obtained from known and unknown TF-gene interactions as described in Section 1.1. On the other hand, there is no concept of training for unsupervised methods.

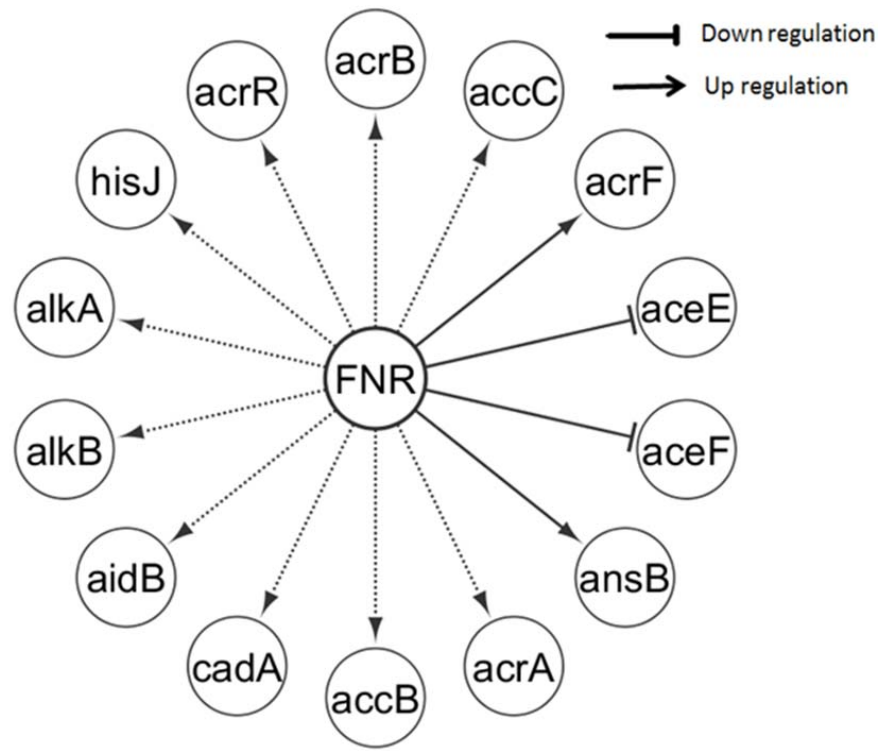

**Figure 1.** Diagram showing the true regulatory relationships between the *E. coli* transcription factor FNR and several target genes. As annotation implies, arrows with solid lines represent up-regulations and inverted T shape solid lines represent down-regulations. These up- and down-regulations together comprise positive examples in this study. Arrows with dotted lines represent unknown interactions, which serve as unlabeled examples in this study.

Specifically, in the supervised methods, the training set contains both positive and negative examples. Known interactions are used as positive examples. However, obtaining negative examples is a challenging task, due to the lack of biological evidence to claim that there is no regulatory connection between a transcription factor and a target gene (Gillani et al., 2014). Some authors (Gillani et al., 2014; Mordelet & Vert, 2008) assumed unknown interactions to be negative examples. However, as explained in Section 1.1, these unknown interactions are not yet verified experimentally; some of them may turn out to be positive examples. For instance, in the first release of RegulonDB (1.0), 533 regulatory interactions were identified. This number was increased to 4,268 in the later release of RegulonDB (8.0), meaning that at the time of the first release 3,735 interactions were unknown, which later turned out to be actually valid interactions (Huerta et al., 1998; Salgado et al., 2013). Under this circumstance, a semi-supervised method is more suitable, which treats all unknown interactions as unlabeled examples and utilizes positive and unlabeled examples to predict gene regulatory networks (GRNs).

The purpose of this study is to investigate semi-supervised methods for GRN prediction. We considered four transcription factors from *E. coli,* namely ARCA*,* CRP*,* FIS and FNR. Similarly we chose four transcription factors from *S. cerevisiae,* namely REB1*,* ABF1*,* CBF1 and GCN4. The four specific transcription factors (TFs) were chosen because they had the largest numbers, ranging from 100 to 400, of known interactions with target genes in the respective organisms. These known interactions were used as positive examples in this study. By utilizing an iterative procedure, we refined the unlabeled examples (i.e., unknown interactions) at hand to get more reliable negative examples for all the four TFs for both organisms.

We adopted inductive and transductive learning approaches for GRN prediction using the semi-supervised methods. With the inductive learning approach, a model is learned from a training dataset, and the model is then applied to a separate testing dataset that is disjoint from the training dataset; hence any information concerning the test data is not seen or used while creating the model (Mitchell, 1997). On the contrary, the transductive learning approach builds a model based on both the training data and some information from the testing dataset, and the model is then applied to the test data (Vapnik, 1998). We employed two machine learning algorithms, namely support vector machines (SVM) and random forests (RF), in this study. The prediction accuracies of both algorithms for the chosen transcription factors of *E. coli* and *S. cerevisiae* were calculated and compared.

## 2. Materials and methods

### 2.1. Datasets

Marbach et al. (2012) performed a comprehensive assessment of network inference methods through the DREAM5 project on both *E. coli* and *S. cerevisiae*. Their study was conducted using two *E. coli* regulatory databases: EcoCyc accessible at http://ecocyc.org/ (Keseler et al., 2013) and RegulonDB accessible at http://regulondb.ccg.unam.mx/ (Salgado et al., 2013). RegulonDB was used in our study because it is a popular database for benchmark experiments. The latest version (version 8.6) of RegulonDB contains 4,268 known TF-gene interactions; these interactions were obtained from the *E. coli* K 12 strand (Salgado et al., 2013). Hence we used the gene expression datasets specifically generated from *E. coli* K 12. These gene expression datasets had GEO accession numbers GSE21869 (Asakura et al., 2011), GSE10158 (Laubacher & Ades, 2008), GSE12411 (Aggarwal & Lee, 2011), GSE33147 (Fong et al., 2005), and GSE17505 (Haddadin & Harcum, 2005). All the datasets are freely available at Gene Expression Omnibus (GEO) (http://www.ncbi.nlm.nih.gov/geo/) and were produced with Affymetrix *E. coli* Antisense Genome Array that contains 7,312 probe sets for *E. coli* gene expression analysis.

For *S. cerevisiae,* there were three regulatory databases used by the DREAM5 study (Marbach et al., 2009, 2010, 2012; Prill et al., 2010), which included YEASTRACT accessible at http://www.yeastract.com/ (Abdulrehman et al., 2011), Fraenkel et al.'s map of conserved regulatory sites accessible at http://fraenkel.mit.edu/improved_map/ (MacIsaac et al., 2006), and the database described in (Hu et al., 2007). The DREAM5 study evaluated these databases and reported that Fraenkel's database contains high quality TF-gene interactions; consequently we used these interactions as positive examples for *S. cerevisiae* in our study. We chose five gene expression datasets for *S. cerevisiae*. These datasets had GEO accession numbers GSE30052 (Chin et al., 2012), GSE12221 (Shalem et al., 2008), GSE12222 (Shalem et al., 2008), GSE40817 (Yona et al., 2012), and GSE8799 (Orlando et al., 2008). All the yeast datasets were created using Affymetrix Yeast Genome 2.0 Array containing 5,744 probe sets for *S. cerevisiae* gene expression analysis.

We extracted expression vectors of TFs and genes that were present in RegulonDB and Fraenkel's database respectively, and created an expression matrix containing the expression vectors for each of the gene expression datasets mentioned above. The *E. coli* expression matrices contained 1,161 gene expressions vectors and the *S. cerevisiae* expression matrices contained 1,994 gene expressions vectors. These matrices were then scaled to zero mean and unit standard deviation.

As explained in Section 1.2, positive examples were created using the known interactions found in RegulonDB and Fraenkel's database for *E. coli* and *S. cerevisiae* respectively. To obtain unknown interactions, we generated all possible combinations of available TF and gene pairs. Each of these combinations was considered as an unknown interaction provided that it did not exist in RegulonDB (Fraenkel's database, respectively) for *E. coli* (*S. cerevisiae*, respectively). These unknown interactions were treated as unlabeled examples. Then all the interactions were separated based on the TFs. For each organism, the top four TFs that had the largest number of interactions were chosen and used to perform the experiments. Table 1 lists the number of positive and unlabeled examples for each chosen TF of *E. coli* and *S. cerevisiae* respectively. The columns named Positive in the table show the total number of known interactions for each TF in *E. coli* and *S. cerevisiae* respectively.

**Table 1.** The number of positive and unlabeled examples for each transcription factor of *E. coli* and *S. cerevisiae* respectively used in this study

| *E. coli* | | | *S. cerevisiae* | | |
|---|---|---|---|---|---|
| TF | Positive | Unlabeled | TF | Positive | Unlabeled |
| CRP | 390 | 770 | REB1 | 217 | 1,776 |
| FNR | 239 | 921 | ABF1 | 199 | 1,794 |
| FIS | 200 | 960 | CBF1 | 164 | 1,829 |
| ARCA | 139 | 1,021 | GCN4 | 120 | 1,873 |

Both supervised and semi-supervised methods work under the principle that if a gene is known to interact with another gene, then any other two genes containing similar gene expression profiles are also likely to interact with each other (Cerulo et al., 2010; Mordelet & Vert, 2008). Based on this principle, feature vectors for TFs and genes were constructed by concatenation of their expression profiles. Hence the resulting feature vectors contained twice the number of features than the original individual expression vectors. While concatenating two profiles, orders were considered, which means, if G1 is known to regulate G2 then the feature vector V (G1, G2) can only be created such that expression values of G1 are followed by expression values of G2. In other words, the feature vector V (G1, G2) implies that G1 regulates G2 but the opposite of that may not necessarily be true. After concatenation, the resulting feature vectors were scaled to zero mean and unit standard deviation.

## 2.2. Proposed semi-supervised methods

Two machine learning algorithms, namely support vector machines (SVM) and random forests (RF), were employed for making predictions. SVM analysis was done using the LIBSVM package in R, accessible at http://cran.r-project.org/web/packages/e1071/index.html (Chang & Lin, 2011). RF analysis was performed using the randomForest package in R, accessible at

http://cran.r-project.org/web/packages/randomForest/index.html (Liaw & Wiener, 2002).

Figure 2 presents the flowchart of the proposed semi-supervised methods for predicting gene regulatory networks (GRNs). Initially we have positive and unlabeled examples, where positive examples represent known TF-gene interactions and unlabeled examples represent unknown interactions. These examples are converted to feature vectors using the method described in Section 2.1, where the feature vectors are used as input of the SVM and RF algorithms. An iterative procedure is then executed; in each iteration a classification model is obtained and validated.

In Figure 2, *Positive* is the set of all positive examples obtained from known interactions of each individual transcription factor of an organism, *Unlabeled* is the set of all unlabeled examples for the organism, $P$ represents the positive training dataset, $N$ represents the negative training dataset, $V$ represents the validation dataset, and $T$ represents the testing dataset. The validation set $V$ contains only positive examples, since true negative examples are not available. All the predictions are made on the examples in $V$, and prediction accuracies of our methods are calculated by comparing true labels with predicted labels. The *Positive* set is evenly divided into two disjoint subsets, $P$ and $V$, where $P$ and $V$ contain approximately the same number of positive examples. For a given transcription factor (TF), $P$ and $V$ remain the same throughout all the iterations for both inductive and transductive learning approaches. $T$ comprises only unlabeled examples, which are used in the iterative procedure to produce reliable negative examples.

Refer to Figure 2. During iteration $k$, $0 \leq k \leq K$, the SVM or RF algorithm is trained using $P$ and $N^k$. A binary classification model, denoted $Model^k$, is obtained. This (SVM or RF) model is then applied to the validation dataset $V$ to predict the labels of the examples in $V$. The prediction accuracy of the model is calculated by dividing the number of correctly predicted examples in $V$ by the total number of examples in $V$. In addition, this model, $Model^k$, is also applied to the testing set $T^k$ to classify the unlabeled examples in $T^k$ as shown in Figure 2. Both SVM and RF algorithms are able to assign probabilistic weights to their classification results. Using these probability values we extract reliable negative examples from the set $T$ by choosing the bottom $|N|$ unlabeled examples in $T$ that have the lowest probability of being positive. These $|N|$ negative examples are collected and stored in $N^{k+1}$, which, together with $P$, will be used to train the SVM or RF algorithm in iteration $k+1$. Notice that iteration 0 is a special case, in which we randomly select $|N|$ examples from the *Unlabeled* set and store them in $N^0$. Notice also that this is a balanced binary classification, since $|P| = |N|$ throughout all the iterations.

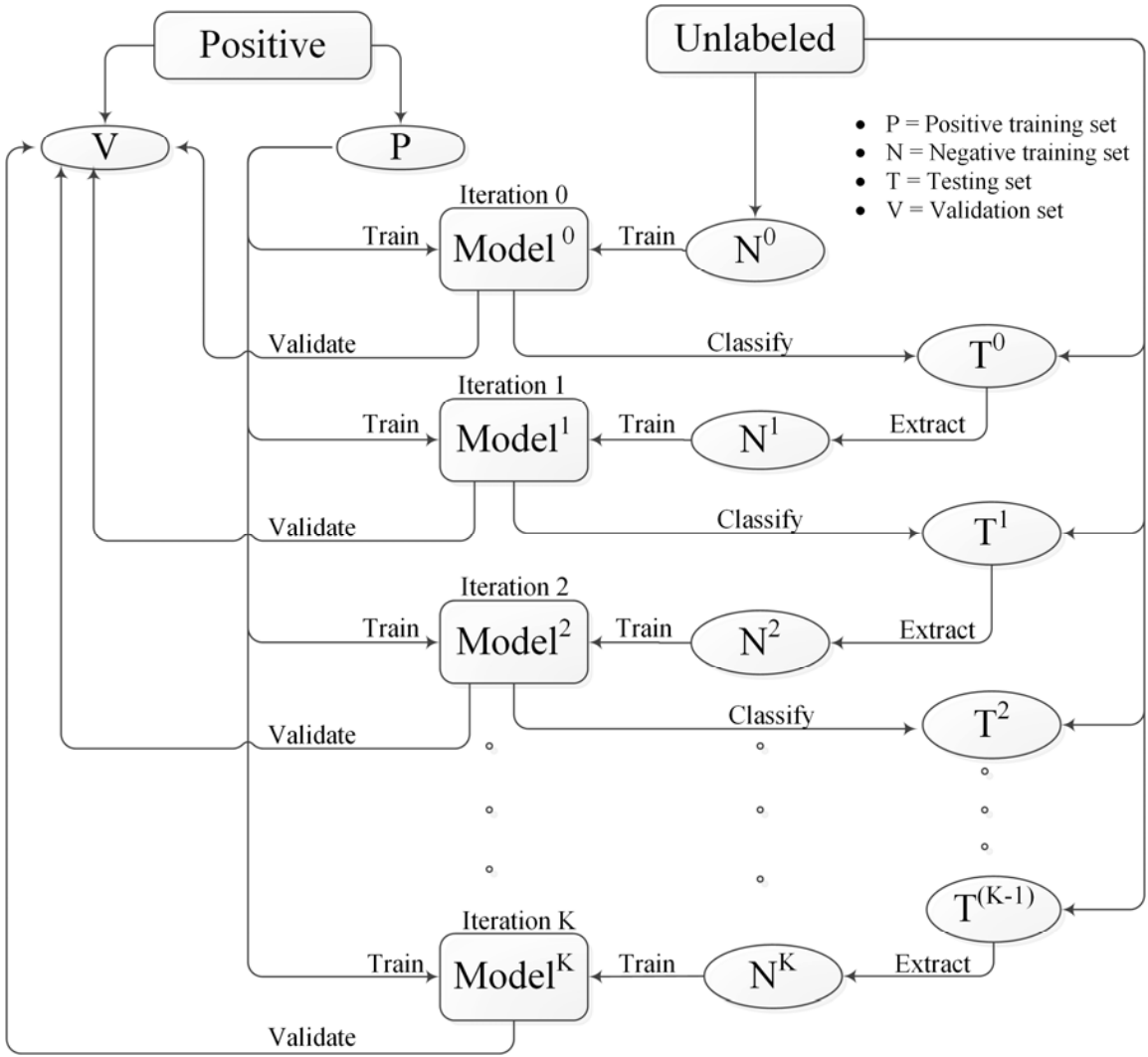

**Figure 2.** Flowchart of the proposed semi-supervised GRN prediction methods.

The difference between the inductive and transductive learning approaches is on whether the sets $N$ and $T$ are disjoint. For the inductive learning approach, $T$ is created such that $N$ and $T$ are disjoint. For the transductive learning approach, however, $T$ includes all the available unlabeled examples (i.e., all the examples in the *Unlabeled* set). More precisely, in iteration $k$, for the inductive learning approach, $T^k = Unlabeled - N^k$ and $N^k \cap T^k = \emptyset$; for the transductive learning approach, $T^k = Unlabeled$ and $N^k \subseteq T^k$. The iterative procedure is executed for each of the four chosen transcription factors (TFs) in each organism. Table 2 lists the number of examples in $P$, $N$, $T$ and $V$ for each TF in *E. coli* and Table 3 shows the data for each TF in *S. cerevisiae.* The Total column in each table shows the total number of training examples, which is equal to $|P| + |N|$, used in each iteration. The Inductive (Transductive, respectively) column in each of these tables shows the number of unlabeled examples in the testing set T used by the inductive (transductive, respectively) learning approach.

**Table 2.** Number of examples in *P*, *N*, *T* and *V* respectively for each TF in *E. coli*

| TF | *P* | *N* | Total | *T* | | *V* |
| --- | --- | --- | --- | --- | --- | --- |
| | | | | Inductive | Transductive | |
| CRP | 195 | 195 | 390 | 575 | 770 | 195 |
| FNR | 120 | 120 | 240 | 801 | 921 | 119 |
| FIS | 100 | 100 | 200 | 860 | 960 | 100 |
| ARCA | 70 | 70 | 140 | 951 | 1021 | 69 |

Table 3. Number of examples in *P*, *N*, *T* and *V* respectively for each TF in *S. cerevisiae*

| TF | P | N | Total | T | | V |
| --- | --- | --- | --- | --- | --- | --- |
| | | | | Inductive | Transductive | |
| REB1 | 108 | 108 | 216 | 1668 | 1776 | 109 |
| ABF1 | 99 | 99 | 198 | 1695 | 1794 | 100 |
| CBF1 | 82 | 82 | 164 | 1747 | 1829 | 82 |
| GCN4 | 60 | 60 | 120 | 1813 | 1873 | 60 |

In all the experiments, the radial basis function (RBF) kernel was used for support vector machines (SVM) with all other parameters set to default values. With random forests (RF), all parameters were set to default values and the number of trees used was 500. For statistical consistency and fair comparisons, the *Positive* and *Unlabeled* sets were kept exactly the same for both SVM and RF in the initial iteration (i.e., iteration 0) for each chosen TF in *E. coli* and *S. cerevisiae* respectively.

## 3. Results

We carried out a series of experiments to evaluate the performance of the proposed semi-supervised methods on the different datasets described in Section 2.1, where the performance of a method was measured by the prediction accuracy of the method. Here, the prediction accuracy of a method is defined as the number of correctly predicted examples in a validation dataset divided by the total number of examples in the validation dataset (cf. Figure 2). Figure 3(a) (Figure 3(b), respectively) compares the transductive and inductive learning approaches with the SVM (RF, respectively) algorithm, where the experimental results were obtained using the *E. coli* transcription factor ARCA and dataset GSE21869. Figure 4(a) (Figure 4(b), respectively) compares the transductive and inductive learning approaches with the SVM (RF, respectively) algorithm, where the experimental results were obtained using the *S. cerevisiae* transcription factor REB1 and dataset GSE12222.

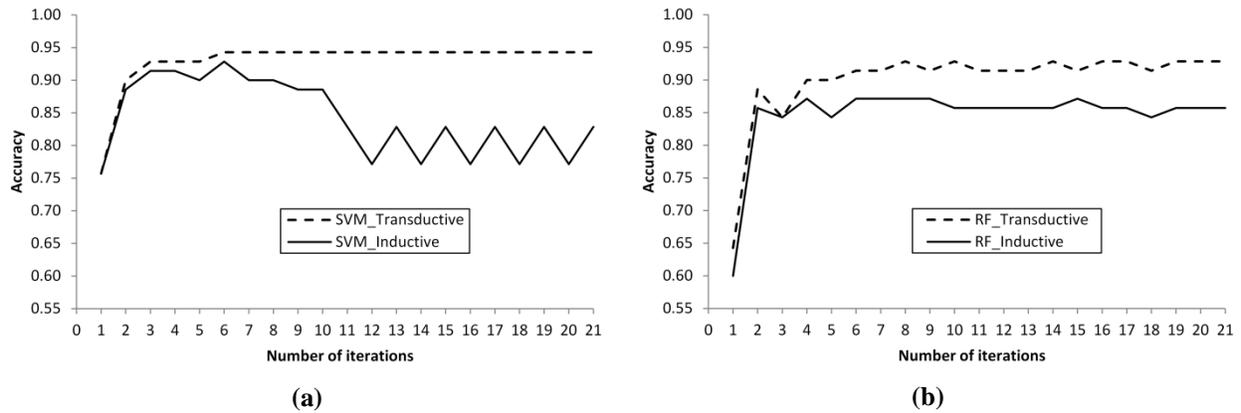

**Figure 3.** Performance comparison of the transductive and inductive learning approaches based on the *E. coli* transcription factor ARCA and dataset GSE21869 with (a) the SVM algorithm, and (b) the RF algorithm.

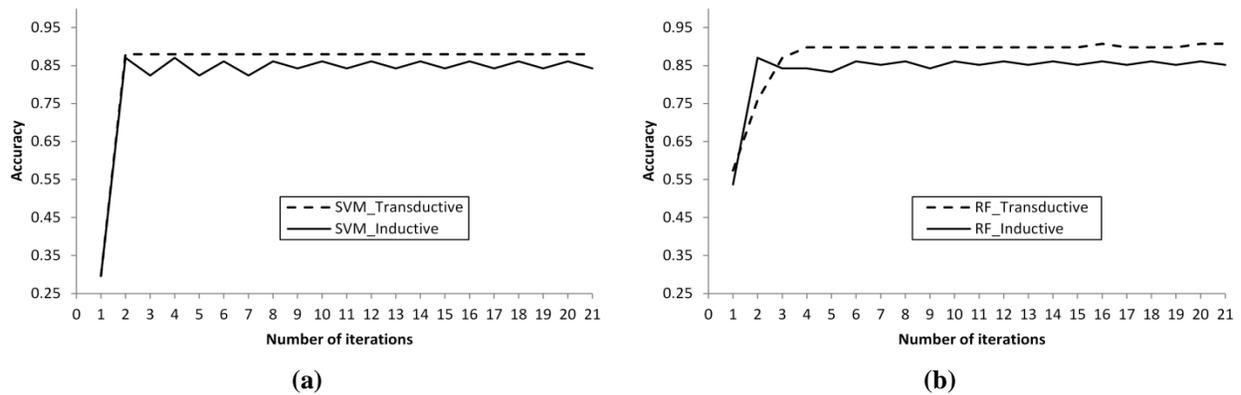

**Figure 4.** Performance comparison of the transductive and inductive learning approaches based on the *S. cerevisiae* transcription factor REB1 and dataset GSE12222 with (a) the SVM algorithm, and (b) the RF algorithm.

It can be seen from Figure 3(a) and Figure 4(a) that, for SVM, the transductive learning approach yielded higher prediction accuracies than the inductive learning approach. The performance of the transductive learning approach became stable as the number of iterations in the proposed semi-supervised methods increased. On the other hand, the performance of the inductive learning approach tended to fluctuate up and down with high frequency. RF exhibited a similar pattern as far as the relative performance of the transductive learning and inductive learning approaches was concerned.

It is worth pointing out that the accuracies of the proposed semi-supervised methods were relatively low when only one iteration (i.e., iteration 0) was executed. Refer to Figure 2. In iteration 0, the negative training set $N^0$ was comprised of randomly selected unlabeled examples. On the other hand, starting from the second iteration (i.e., iteration 1), the proposed semi-supervised methods picked unlabeled examples that had the lowest probability of being positive and used them as negative training examples. These unlabeled examples with the lowest probability of being positive formed more reliable negative training data than the randomly selected unlabeled examples, hence yielding higher accuracies.

In subsequent experiments, we adopted the transductive learning approach and fixed the number of iterations at 15. We applied our semi-supervised GRN prediction methods, with both SVM and RF, to the different gene expression datasets for the different transcription factors selected from *E. coli* and *S. cerevisiae* respectively. Figure 5(a) (5(b), 5(c), 5(d) respectively) shows a bar chart comparing the performance of SVM and RF using the gene expression datasets GSE10158, GSE12411, GSE33147, GSE21869 and GSE17505 for the *E. coli* transcription factor ARCA (CRP*,* FIS*,* FNR respectively). Figure 6(a) (6(b), 6(c), 6(d) respectively) shows a bar chart comparing the performance of SVM and RF using the gene expression datasets GSE30052, GSE12221, GSE12222, GSE40817 and GSE8799 for the *S. cerevisiae* transcription factor REB1 (ABF1*,* CBF1*,* GCN4 respectively).

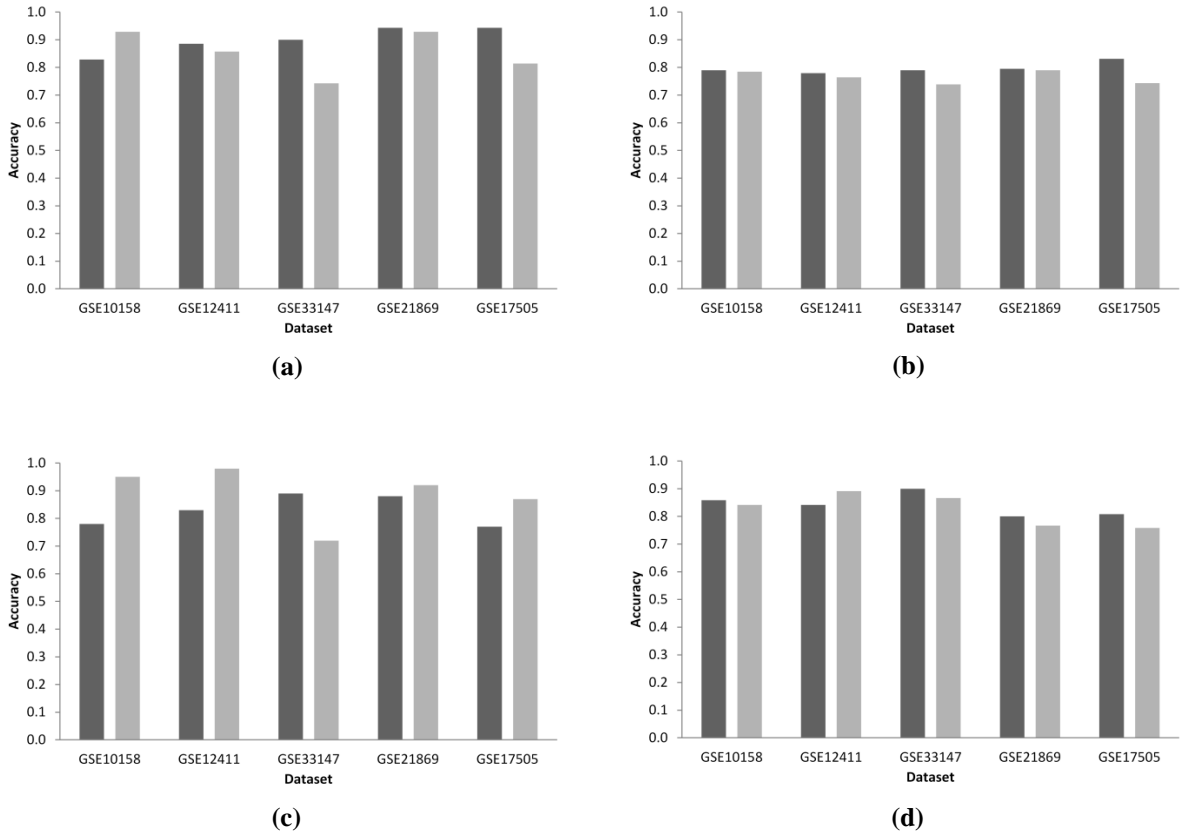

**Figure 5.** Performance comparison of the SVM and RF algorithms with the transductive learning approach on five gene expression datasets GSE10158, GSE12411, GSE33147, GSE21869 and GSE17505, and four transcription factors of *E. coli* including (a) ARCA, (b) CRP, (c) FIS, and (d) FNR. Dark bars represent SVM and light bars represent RF.

It can be seen from Figure 5(a) that SVM yielded higher prediction accuracies than RF on the datasets GSE33147 and GSE17505 for the *E. coli* transcription factor ARCA while RF performed better than SVM on the dataset GSE10158. The two machine learning algorithms exhibited similar performance on the datasets GSE12411 and GSE21869. Figure 5(b) shows that SVM and RF exhibited similar performance on all datasets except GSE17505, where SVM was more accurate than RF for the transcription factor CRP. Significant discrepancies were observed in the predictions accuracies with the transcription factor FIS, where RF outperformed SVM on all datasets except GSE33147 (Figure5(c)). Finally, as Figure 5(d) implies, SVM and RF did not show any major difference in GRN prediction for the *E. coli* transcription factor FNR.

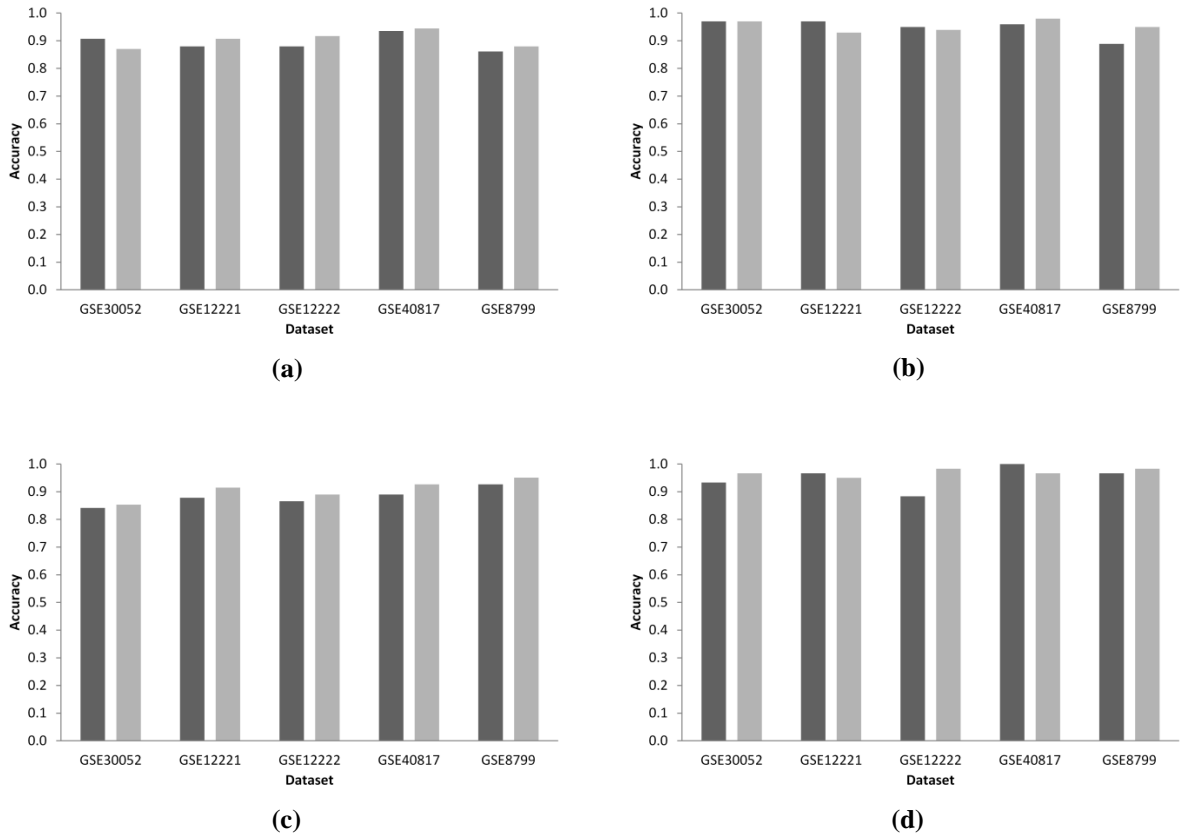

**Figure 6.** Performance comparison of the SVM and RF algorithms with the transductive learning approach on five gene expression datasets GSE30052, GSE12221, GSE12222, GSE40817 and GSE8799, and four transcription factors of *S. cerevisiae* including (a) REB1, (b) ABF1, (c) CBF1, and (d) GCN4. Dark bars represent SVM and light bars represent RF.

Figure 6 shows that the performance of SVM and RF was nearly identical across all the datasets for all the *S. cerevisiae* transcription factors used in this study. There were only two instances where significant differences were observed. The first instance was on the dataset GSE8799 for the *S. cerevisiae* transcription factor ABF1 (Figure 6(b)). The second instance was on the dataset GSE12222 for the transcription factor GCN4 (Figure 6(d)). In both instances, RF performed better than SVM.

# 4. Discussion

For each of the organisms *E. coli* and *S. cerevisiae,* we chose four transcription factors having enough known TF-gene interactions, and evaluated the effectiveness of our proposed semi-supervised methods. These methods employed an iterative procedure together with a transductive learning strategy or an inductive learning strategy to obtain more reliable negative training examples. Our experimental results indicated that the transductive learning approach consistently outperformed the inductive learning approach on the datasets tested in this study. The results also showed that after certain iterations, the prediction accuracy of the transductive learning approach tended to converge. For many experiments the convergent point was identified within the first 10 interactions. To provide more reliable and consistent findings we therefore fixed the number of iterations at 15 in performing the rest of the experiments. With 15 iterations, no significant difference in prediction accuracies was observed between the SVM and RF algorithms using the transductive learning approach.

On the other hand, the inductive leaning approach did not exhibit a clear convergent point; instead, it exhibited waving patterns; cf. Figure 3. Such a behavior might exist due to the fact that in the inductive learning approach a portion of TF-gene interactions were not used. As explained in Section 2.2, in iteration $k$, $T^k = Unlabeled - N^k$, and we extract some unlabeled examples from $T^k$ to get negative training examples in $N^{k+1}$ that are used in iteration $k+1$. $T^k$ does not contain the TF-gene interactions in $N^k$. Hence when the TF-gene interactions in $T^k$ are ranked in the descending order of their probabilities of being positive, they do not contain the TF-gene interactions in $N^k$, and therefore these TF-gene interactions in $N^k$ will not be in $N^{k+1}$. It is likely that $N^k$ may contain some TF-gene interactions that are very reliable negative examples. As a consequence, these very reliable negative examples will not be in $N^{k+1}$ used for training the machine learning algorithms (SVM or RF) in iteration $k+1$. Hence when those very reliable negative examples are in the training set, the trained model yields a high accuracy; otherwise the trained model yields a low accuracy. This explains why the inductive learning approach exhibited waving patterns in terms of prediction accuracies. Note that, in the transductive learning approach, $T^k$ = *Unlabeled*, and hence those very reliable negative examples are always considered in every iteration. As a consequence, the performance of the transductive learning approach became stable when sufficient high-quality negative examples were collected after a certain number of iterations (e.g., after 15 iterations).

The experimental results in Section 3 were obtained using the radial basis function (RBF) kernel for the SVM algorithm and 500 trees for the RF algorithm. We also tested the linear kernel and polynomial kernel available in the LIBSVM package with default parameter values on all five gene expression datasets and for all four transcription factors. For the RF algorithm, we tested it using different numbers of trees, specifically 100, 500 and 1,000 trees respectively. The number of iterations used in the semi-supervised methods was fixed at 15. The results obtained were similar to those presented in Section 3. These results indicate that all the three kernels, namely RBF, linear and polynomial kernels, work well. Furthermore, the number of decision trees used in the RF algorithm has little impact on its performance provided the number is sufficiently large (e.g., at least 100). However, with too many trees (e.g., the number of trees is greater than 1,000), the time of the RF algorithm may increase substantially since it takes a large amount of time to build these trees.

Refer to Figures 3 and 4. The performance of the SVM algorithm clearly converged in both organisms with the transductive learning strategy in the sense that after a certain number of iterations (e.g., after 15 iterations), the accuracies of the algorithm did not vary too much. The reason behind this phenomenon is that, after 15 iterations, the algorithm has identified the most reliable negative training examples, which remain the same for subsequent iterations. Therefore the models created for those subsequent iterations by the reliable negative training examples and the positive training set $P$ remain almost the same, and hence always make the same predictions; cf. Figure 3(a) and Figure 4(a).

On the other hand, referring to Figure 3(b) and Figure 4(b), we see that, with the transductive learning strategy, there were slight variations in the performance of the RF algorithm even after 15 iterations were executed. Although those variations are negligible, allowing us to draw the qualitative conclusions based on our findings, a close look at the machine learning algorithms explains why the variations occur. In general, the SVM algorithm systematically attempts to find a hyperplane that maximizes the distance to the nearest training example of any class (Joachims, 1999). There is no randomness associated with the SVM algorithm. On the contrary, the RF algorithm randomly picks training examples and features in the training examples to build decision trees (Breiman, 2001). Due to the randomness associated with the RF algorithm, the strong convergence was not observed for the algorithm; cf. Figure 3(b) and Figure 4(b).

It is worth pointing out that the proposed semi-supervised methods performed better than the supervised methods described in (Gillani et al., 2014; Mordelet & Vert, 2008). Just like how the semi-supervised methods work in the first iteration (i.e., iteration 0) where randomly chosen unlabeled examples in $N^0$ and positive examples in $P$ are used to train a machine learning algorithm (e.g., SVM), the supervised methods treat the unlabeled examples as negative examples and use them together with the positive examples in $P$ to train the SVM algorithm (Gillani et al., 2014; Mordelet & Vert, 2008). As shown in Figures 3 and 4, executing merely the first iteration without iteratively refining the unlabeled examples to obtain more reliable negative training examples performs worse than executing several (e.g., 15) iterations as done by the proposed semi-supervised methods, suggesting that the proposed semi-supervised methods be better than the supervised methods.

## 5. Conclusions

The idea of training a classifier using positive and unlabeled examples was previously proposed to classify web pages (Blum & Mitchell, 1998) and text documents (Li & Liu, 2003; Liu et al., 2003). We extend this idea here to predict gene regulatory networks using both inductive and transductive learning approaches. To utilize available unlabeled examples and to effectively extract reliable negative examples, we adopted support vector machines (SVM) and random forests (RF), both of which were able to assign probabilistic weights to their classification results. We picked negative examples from the testing set that had the lowest probability of being positive. Our experimental results showed that the negative examples chosen this way yielded better performance than the negative examples that were randomly selected from the unlabeled data.

In addition, our results showed that the transductive learning approach outperformed the inductive learning approach and exhibited a relatively stable behavior for both SVM and RF algorithms on the datasets used in this study. Algorithmic parameters such as different kernels for SVM and different numbers of decision trees for RF did not yield significantly different results. Furthermore, there was no clear difference in the performance of SVM and RF for both the prokaryotic organism (*E. coli*) and the eukaryotic organism (*S. cerevisiae*). The experimental results also showed that our proposed semi-supervised methods were more accurate than the supervised methods described in (Gillani et al., 2014; Mordelet & Vert, 2008) since the semi-supervised methods adopted an iterative procedure to get more reliable negative training examples than those used by the supervised methods.

A shortcoming of our semi-supervised methods is that, sometimes not enough known TF-gene interactions are available for certain transcription factors, or there exists no known TF-gene interaction at all for some organisms. Under this circumstance, semi-supervised methods like what we describe here may yield low prediction accuracies due to the lack of reliable training data. One possible way for dealing with organisms with only unknown TF-gene interactions or unlabeled examples is to use SVM and RF algorithms to assign probabilistic weights to their classification results. Then pick positive (negative, respectively) examples from the testing set that have the highest (lowest, respectively) probability of being positive, and use these more reliable positive and negative training data to obtain a hopefully better classification model. In future work, we plan to investigate the performance of this approach and compare different machine learning algorithms using this approach.

Another line of future work is to compare semi-supervised and supervised methods using simulated data such as those available in the DREAM4 project (Marbach et al., 2009, 2010, 2012). Preliminary analysis on the simulated data indicated that the two methods are comparable, though both can be improved by adopting more features in addition to the gene expression profiles used here. Further research will be conducted to develop additional biologically meaningful features and to evaluate the effectiveness of those features.